\documentclass{article}
\usepackage{amsmath}
\usepackage{algpseudocode,algorithm}
\PassOptionsToPackage{compress,numbers}{natbib}
\usepackage[preprint]{neurips_data_2022}

\usepackage[utf8]{inputenc} 
\usepackage[T1]{fontenc}    
\usepackage{hyperref}       
\usepackage{url}            
\usepackage{booktabs}       
\usepackage{amsfonts}       
\usepackage{nicefrac}       
\usepackage{microtype}      
\usepackage{xcolor}         
\usepackage{comment}
\usepackage{float}
\usepackage{graphicx}
\usepackage{bbm}
\usepackage{enumitem}
\usepackage{verbatim}

\newcommand{\id}{\ensuremath{\mathbbm{1}}}
\newcommand{\sh}{\text{shape}}
\newcommand{\she}{\text{shape}^{\emptyset}}
\newcommand{\co}{\text{color}}
\newcommand{\coe}{\text{color}^{\emptyset}}
\newcommand{\bu}{\text{bucket}}
\newcommand{\bue}{\text{bucket}^{\emptyset}}
\newcommand{\x}{\times}

\title{The Game of Hidden Rules: A New Kind of Benchmark Challenge for Machine Learning}

\author{Eric Pulick$^{1*}$, Shubham Bharti$^1$, Yiding Chen$^1$, Vladimir Menkov$^2$\\ \textbf{Yonatan Mintz$^1$, Paul Kantor$^1$, Vicki M.~Bier$^1$}\\
 $^1$University of Wisconsin - Madison, $^2$Rutgers University\\
\{pulick, sbharti, ychen695, ymintz, pkantor, vmbier\}@wisc.edu, vmenkov@gmail.com
}
\begin{document}

\maketitle

\begin{abstract}
As machine learning (ML) is more tightly woven into society, it is imperative that we better characterize ML's strengths and limitations if we are to employ it responsibly. Existing benchmark environments for ML, such as board and video games, offer well-defined benchmarks for progress, but constituent tasks are often complex, and it is frequently unclear how task characteristics contribute to overall difficulty for the machine learner. Likewise, without a systematic assessment of how task characteristics influence difficulty, it is challenging to draw meaningful connections between performance in different benchmark environments. We introduce a novel benchmark environment that offers an enormous range of ML challenges and enables precise examination of how task elements influence practical difficulty. The tool frames learning tasks as a ``board-clearing game,'' which we call the Game of Hidden Rules (GOHR). The environment comprises an expressive rule language and a captive server environment that can be installed locally. We propose a set of benchmark rule-learning tasks and plan to support a performance leader-board for researchers interested in attempting to learn our rules. GOHR complements existing environments by allowing fine, controlled modifications to tasks, enabling experimenters to better understand how each facet of a given learning task contributes to its practical difficulty for an arbitrary ML algorithm.
\end{abstract}

\section{Introduction}\label{sec:Int}
Learning computational representations of rules has been one of the main objectives of the field of machine learning (ML) since its inception. In contrast to pattern recognition and classification (the other main domains of ML), rule learning is concerned with identifying a policy or computational representation of the hidden process by which data has been generated. These sorts of learning tasks have been common in applications of ML to real world settings such as biological research \citep{khatib_algorithm_2011}, imitation learning  \citep{hussein_imitation_2018}, and most recently game play \citep{mnih_human-level_2015,silver_general_2018}. Since this process involves sequential experimentation with the system, much of the recent work exploring rule learning has focused on using reinforcement learning (RL) for learning rules as optimal policies of Markov decision processes.

Rules, however, can be quite complex in many applications, and may not map to a single optimal policy, but instead to a whole set of optimal policies. For instance, in the case of protein folding, there may be multiple correct ways in which a potential set of atoms can be folded into a feasible molecule. While computational and theoretical results exist to characterize which properties of the models or algorithms may cause these differences, these results are highly task-dependent, making it difficult to compare relative task difficulty. Thus, an important question is whether some characteristics make particular rules easier or harder to learn by a specific algorithm (or in general). To date, this has been a difficult question to answer, since many rules of interest in the real world are multifaceted and not well characterized. For instance, while there are effective RL algorithms that can play go, chess, and backgammon, it is not clear which specific features of go and chess make them more difficult to learn in practice than backgammon (e.g., is it the number of pieces? size of the board? lack of random elements?). In order to interrogate these questions, new ways of generating rules and data must be devised that allow for researchers to examine these characteristics in a controlled environment.

In this paper, we propose a new data environment called the Game Of Hidden Rules (GOHR), which aims to help researchers in this endeavor. The main component of the environment is a game played in a $6\times 6$ board with game pieces of different shapes and colors. The task of the learner is to clear the board in each round by moving the game pieces to ``buckets'' at the corners of the board according to a hidden rule, known to the researcher but not to the learner. Our environment allows researchers to express a hidden rule using a rich syntax that can map to many current tasks of interest in both the classification and RL settings. A key advantage of our environment is that researchers can control each aspect of the hidden rules and test them at a granular level, allowing for experiments that determine exactly which characteristics make some learning tasks harder than others and which algorithms are better at learning specific types of rules. 

The rest of the paper proceeds as follows. We first describe how our environment relates to other data-generating environments from the literature in Section~\ref{sec:lit_rev}. Then, in Section~\ref{sec:gohr}, we go into the details of GOHR and the rule syntax, explaining how the environment can be used to interrogate the effects of rule structure. We introduce our ML competition and refer readers to benchmark rules, instructions, and documentation available at our \href{http://sapir.psych.wisc.edu:7150/w2020/captive.html}{public site} in Section~\ref{sec:doc-comp}. In Section~\ref{sec:sample}, we present sample rules and analysis for an example algorithm. Finally, we conclude with some discussion on the implications of our results and on other questions that can be answered by the GOHR environment in Section \ref{sec:discussion}.

\section{Literature Review}
\label{sec:lit_rev}

Learning rules remains a core challenge in many aspects of ML, particularly in RL settings where the agent must learn iteratively from its environment. Games have historically served as rich benchmark environments for RL, with novel challenges in each game environment spurring progress for the field as a whole. RL has tackled increasingly complex classical board games, such as backgammon, chess, shogi, and go \citep{tesauro_td-gammon_1994,campbell_deep_2002, silver_mastering_2016, silver_mastering_2017, silver_general_2018}, eventually surpassing the performance of human experts in each. Of late, video-game environments have also become drivers of progress in RL. Beginning with human-level performance in Atari 2600 games \citep{mnih_human-level_2015,badia_agent57_2020}, machine players have become competitive with humans in a variety of environments, including real-time-strategy and multiplayer games such as Quake III, StarCraft II, and Dota 2 \citep{jaderberg_human-level_2019, openai_dota_2019, vinyals_starcraft_2017, vinyals_grandmaster_2019}. 

Instrumental in such RL progress has been the growing space of benchmark game environments and supporting tools, such as the Arcade Learning Environment \citep{bellemare_arcade_2013}, General Video Game AI \citep{perez-liebana_2014_2016}, OpenAI Gym \citep{brockman_openai_2016}, Gym Retro \citep{nichol_gotta_2018}, Obstacle Tower \citep{juliani_obstacle_2019}, Procgen \citep{cobbe_leveraging_2020}, and NetHack environments \citep{kuttler_nethack_2020, samvelyan_minihack_2021}. Taken together, these represent an impressive range of new tasks for RL agents, bridging many gaps and challenges in achieving aspects of artificial intelligence. 

We note a clear trend in these benchmark tools away from static games and toward configurability (e.g. procedurally-generated and customizable environments), something we believe is critical to better understanding task difficulty. Many environments demand that agents solve complex tasks inspired by those a human might encounter in the real world. Such tasks offer well-defined benchmarks, but often do not support precise, controlled modifications without affecting other aspects of the learner's environment. We believe that GOHR, which allows the experimenter to change the learning task without any change to the game's rules, playing environment, or board-clearing objective, can serve as an important complement to existing benchmarks. GOHR's mechanics allow for isolated study of changes in learning tasks and their effects on practical difficulty, an important prerequisite for the systematic study of ML algorithms and their suitability for different tasks.

\section{Game of Hidden Rules}
\label{sec:gohr}
In this section we describe the GOHR's structure, syntax, and the expressivity of the rule language. 
In each episode of our game, the player is presented with a game board containing a configurable set of game pieces, each drawn from a standard set of shapes and colors. The player's objective is to remove game pieces from play by dragging and dropping them into buckets located at each corner of the game board. A ``hidden rule,'' unknown to the player, determines which pieces may be placed into which buckets at a given point in the game play. For instance, a rule might assign game pieces to specific buckets based on their shape or color. If the player makes a move permitted by the rule, the corresponding game piece is removed from play; otherwise, it remains in its original location. The episode concludes once the player has fully satisfied the rule. Typically, this occurs when all game pieces have been cleared from play, but some rules may be fully satisfied even with some pieces remaining on the board. The scoring for GOHR rewards players for completing episodes in as few moves as possible, incentivizing players to quickly discern the hidden rule.

\begin{figure}[h]
    \centering
    \begin{minipage}{.45\textwidth}
        \centering
        \includegraphics[width=.9\textwidth]{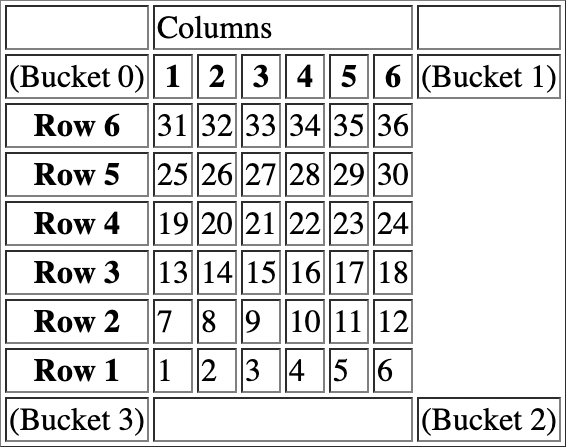}
        \caption{Game board diagram, with numeric position labels included.}
        \label{fig:plain-board}
    \end{minipage}
    \hspace{1cm}
    \begin{minipage}{0.45\textwidth}
        \centering
        \includegraphics[width=.7\textwidth]{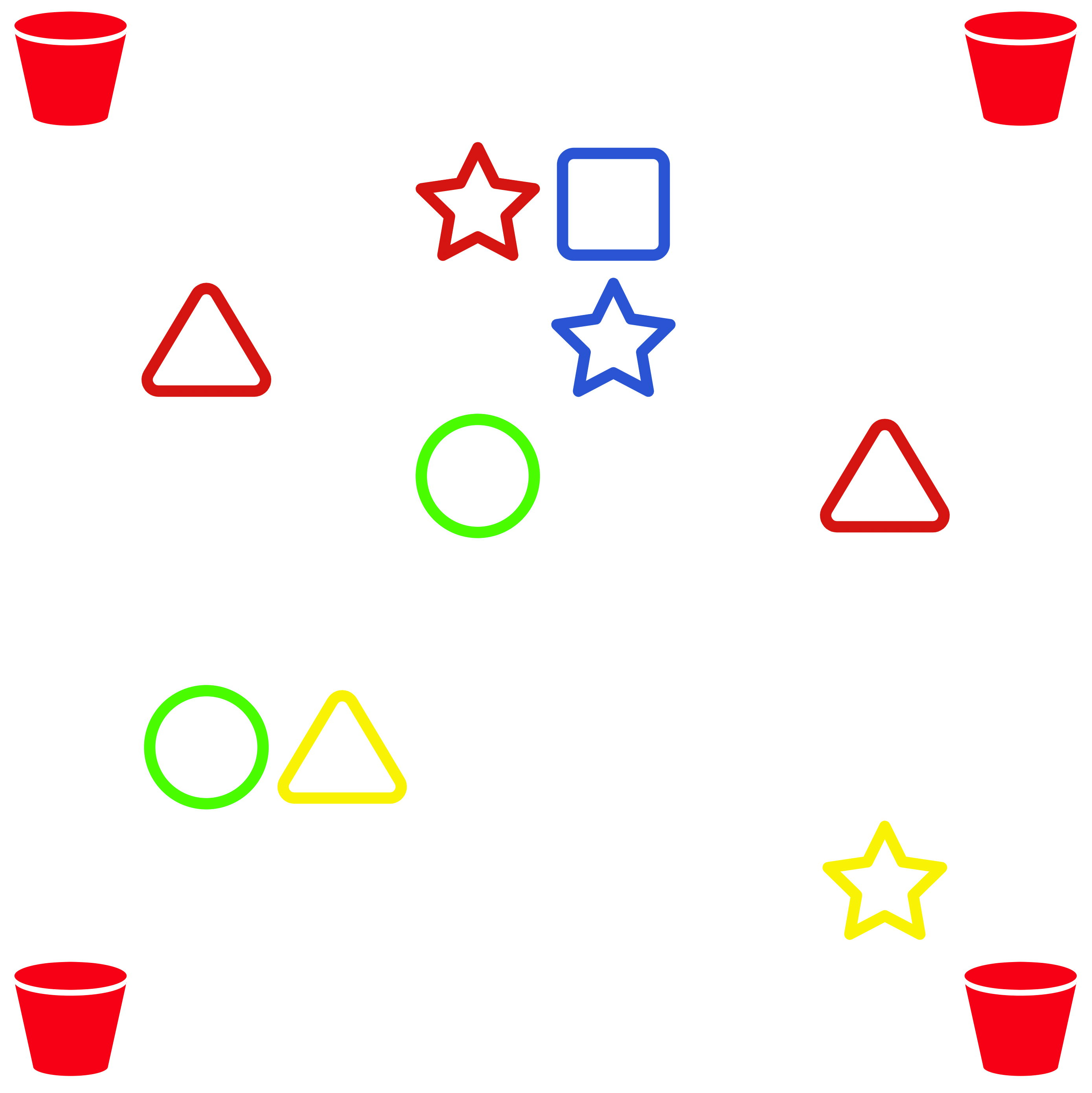}
        \caption{Game board, configured with 4 colors and 4 shapes, as it would be seen by a human player.}
        \label{fig:actual-board}
    \end{minipage}
\end{figure}

The GOHR is played on a board with 36 identical cells, arranged in a $6\times6$ grid. Each cell on the board is indexed with a label, 1-36, and has $x$ and $y$ coordinates 1-6. Each bucket is indexed with a label, 0-3, with $x,y$ coordinates in the set $\{(0,0),(0,7),(7,0),(7,7)\}$. The game board, including all numeric labels, can be seen in Figure~\ref{fig:plain-board}.

In each GOHR experiment, the game engine generates game pieces from user-defined sets of shapes and colors. Depending on the experimental goals, these sets can be of arbitrary size. If no specification is provided in the experimental setup, the game engine defaults to a set of 4 shapes (circles, triangles, squares, stars) and 4 colors (red, blue, black, yellow). The experiment designer may add shapes or colors to the experiment in the associated color and shape configuration files. A sample board, built using a set of 4 shapes and 4 colors, is shown in Figure~\ref{fig:actual-board}.

Boards in the GOHR can be defined in advance of play (to permit precise experimenter control), or can be generated randomly by the game engine according to a set of input parameters. This flexibility allows the experimenter to evaluate the impact of deterministic or randomized ``curricula'' (e.g., to determine whether seeing a particular set of boards in a specified order is conducive to learning).
When generating boards randomly, the experimenter must specify the minimum and maximum numbers of game pieces, colors, and shapes to appear on each new board; the game engine randomly selects values in the range [minimum, maximum] for each quantity and generates boards accordingly. Additional documentation can be found at our \href{http://sapir.psych.wisc.edu:7150/w2020/captive.html}{public site}, as noted in Section~\ref{sec:doc-comp}.

\subsection{Rule definition}
A rule playable within GOHR must be defined in a standalone file. Each rule is constructed from one or more \textbf{rule lines}, each of which is built from one or more \textbf{atoms}. For instance, a two-line rule with five atoms might look like:
\begin{center}
    (atom 1) (atom 2) (atom 3)\\
    (atom 4) (atom 5)
\end{center}

Only one rule line is \textbf{active} at a time;  this active line determines the current \textbf{rule state} (how game pieces may be placed into buckets for the player's next move). In the example above, the rule state is formed by the contents of either atoms 1, 2, and 3 or atoms 4 and 5, depending on which line is active. Later sections discuss the mechanisms by which the active rule line can change during play.

\begin{samepage}Each atom maps a set of game pieces to a set of permitted buckets and is defined as follows:
\begin{center}
    (count, shapes, colors, positions, buckets)
\end{center}
\end{samepage}
The ``count'' parameter defines the number of successful moves that the atom permits (also referred to as its ``metering''). Any game pieces matching the ``shapes,'' ``colors,'' and ``positions'' specified in the atom are accepted in its corresponding set of ``buckets.'' Multiple values can be listed for each non-count field, and are grouped in square brackets. For example, the rule that ``stars and triangles go in bucket 0, squares go in bucket 1, and circles can go in bucket 2 and bucket 3'' would be constructed with the following three atoms on a single rule line:
\begin{center}
    (*, [star, triangle], *, *, 0) (*, square, *, *, 1) (*, circle, *, *, [2,3])
\end{center}
The $*$ character denotes a wildcard. In the count field, $*$ denotes that the atom is not metered; in the shape, color, or position fields, it denotes that every value for that feature is included in the allowable set for that atom. In the above example, the atoms apply to any star, triangle, square, or circle on the board, regardless of their color or position. Since the atoms are not metered, any shape may be placed in its corresponding bucket at any point in the game.

\subsection{Rule expressivity}
The above syntax permits an enormous range of rules within the GOHR. \textbf{Stationary} rules, such as the previous example, are those where the permitted bucket set for each game piece never changes during play. In that example, any star or triangle is allowed only in bucket 0, any square allowed only in bucket 1, etc., for as long as the game is played. Thus, stationary rules permit the experimenter to study the practical difficulty associated with learning feature-based patterns--i.e., those related to game-piece color, shape, position, or a combination thereof. Though stationary rules have no dependence on play history, these rules can still be made arbitrarily difficult, such as designating a specific bucket for every possible combination of a game piece's features. Importantly, since the set of colors and shapes used is left to the experimenter, such rules can become arbitrarily complex.

\textbf{Non-stationary} rules are those in which the permitted bucket set for game pieces changes during play. Non-stationary rules can involve following a defined sequence, such as as clearing a triangle, a square, a circle, and then a star in that order (repeating the sequence as needed). It is also possible to create non-stationary rules with an implicit priority between objects, such as clearing all available red pieces before all available blue pieces, or clearing pieces on the board from top to bottom. Methods for implementing non-stationary rules are described below.

As mentioned, each atom contains a ``count'' field dictating the number of corresponding correct moves that atom permits. For example, the following rule requires the player to alternate between placing game pieces in the top (0,1) and bottom (2,3) buckets:
\begin{center}
    (1, *, *, *, [0,1])\\
    (1, *, *, *, [2,3])
\end{center}
The first rule line is always active at the beginning of a new episode, in this case allowing any game piece on the board to be placed into either bucket 0 or bucket 1. Once a player makes a move satisfying one (or more) of the atoms on the active rule line, the counts associated with the satisfied atoms are decremented by one. An atom with a count of 0 is considered exhausted and no longer accepts moves. At each move, the game engine evaluates the active rule line and the current board; if there are no valid moves available to the player, the engine makes the next rule line active and resets all counts in the new line. Thus, when all atoms on the active rule line are exhausted, there are no valid moves available and the engine moves control to the next rule line. If there are no lines below the current rule line, the first rule line is made active again. In this example, when the player puts a game piece in buckets 0 or 1, the first atom becomes exhausted and the second line becomes active. After placing a piece into either bucket 2 or 3, the second atom is exhausted and the first line is reset and becomes active again, repeating this cycle until the board is cleared.

GOHR also permits the experimenter to attach a count to each rule line. When the line is active, this count decrements each time any atom on the line is satisfied. For instance, a rule that alternates between uniquely assigning shapes and colors to buckets would look like:
\begin{center}
    1 (*, star, *, *, 0) (*, square, *, *, 1) (*, circle, *, *, 2) (*, triangle, *, *, 3)\\
    1 (*, *, red, *, 0) (*, *, blue, *, 1) (*, *, black, *, 2) (*, *, yellow, *, 3)
\end{center}
If the rule-line count is exhausted, the game engine makes the next line active, even if there are non-exhausted atoms on that line. For this example, the active rule line will alternate after each successful move, regardless of which atom in the active line is satisfied. If no count is provided for a given rule line, the game engine assumes that the rule line is not metered and can be used until all atoms on that line are exhausted or no valid move exists for the game pieces currently on the board.

The GOHR rule syntax allows the experimenter to write expressions in an atom's bucket field that define sequences based on previous successful moves. The game engine stores values for the bucket that most recently accepted any game piece ($p$) as well as the bucket that most recently accepted an item of each color ($pc$) and shape ($ps$). A simple rule expressible using these values is ``objects must be placed in buckets in a clockwise pattern, beginning with any bucket'':

\begin{center}
    (1, *, *, *, [0,1,2,3]) \\
    (*, *, *, *, p+1)
\end{center}

The expressions used in the buckets field are evaluated modulo 4 to ensure that the resultant expression gives values in the range 0-3. The game engine also supports the terms ``nearby'' and ``remotest'' as bucket definitions, which allow the experimenter to require that game pieces be put into either the closest or furthest bucket, respectively, evaluated by Euclidean distance. 

The arrangement of atoms allows the experimenter to encode priority between component tasks within a rule. For instance, the rule that ``all red pieces go into bucket 1, \emph{then} all blue pieces go into bucket 2'' would be expressed as follows:
\begin{center}
    (*, *, red, *, 1) \\
    (*, *, blue, *, 2)
\end{center}
Since both the first rule line and associated atom are not metered, they can never become exhausted, even if there are no more red game pieces left on the board. However, as noted above, the engine evaluates if there are any valid moves available to the player given the current rule line; if there are no valid moves available, the engine shifts control to the next line. In this example, once the player has cleared all red game pieces from the board, the engine will make the second rule line active.

\section{Documentation and Competition} \label{sec:doc-comp}
Beyond using GOHR for the study of task characteristics and their impact on difficulty, we hope that interested researchers will share performance results on benchmark rules with us so that we can disseminate community-wide benchmark performance. To this effect, we plan to host a public leader-board through June 2023 which will present results for benchmark rules, updated on a weekly basis. The public site, \url{http://sapir.psych.wisc.edu:7150/w2020/captive.html}, houses documentation for downloading and setting up the captive game server (CGS), configuring and running experiments, a list of benchmark rules beyond what is presented in Section~\ref{sec:sample}, as well as the public leader-board.

\section{Sample Analysis} \label{sec:sample}
In this section we propose a series of sample rules, a sample ML algorithm (MLA), performance metrics, and useful data presentation methods.

\subsection{Sample rules}
We present results for the following sample rules, each played using the default set of 4 shapes and colors. Boards were randomly generated with 9 game pieces per board and the minimum and maximum parameters for color and shape set to 4.
\begin{enumerate}
    \item \textbf{Color Match}: Each of the 4 shapes is uniquely assigned to a bucket
        \subitem (*, star, *, *, 0) (*, triangle, *, *, 1) (*, square, *, *, 2) (*, circle, *, *, 3)
    \item \textbf{Clockwise}: The first game piece can go in any bucket, but each subsequent piece must go in the next clockwise bucket
        \vadjust{\penalty10000}
        \subitem (1, *, *, *, [0,1,2,3])
        \vadjust{\penalty10000}
        \subitem (*, *, *, *, (p + 1))
    \item \textbf{B23 then B01}: Game pieces must be alternatively placed in the bottom and top buckets
        \subitem (1, *, *, *, [2,3])
        \subitem (1, *, *, *, [0,1])
    \item \textbf{B3 then B1}: Game pieces must be alternatively placed in buckets 3 and 1
        \subitem (1, *, *, *, 3)
        \subitem (1, *, *, *, 1)
\end{enumerate}

\subsection{Sample MLA}
We model the GOHR as an episodic Markov decision process (MDP) characterized by the tuple, $\mathcal M = (\mathcal S, \mathcal A, R, P, \mu, H)$.
$\mathcal A$ is the action space, where each action puts the object in row $r$ and column $c$ in bucket $b$: 
$\mathcal A= \{(r, c, b) : r \in [6], c \in [6],  b \in \{(0,0), (0,7), (7,0), (7,7)\} \}$.
Before introducing the state space, we first formally describe the board.
In a board $B$, the cell at row $r$, column $c$ is characterized by color and shape:
$B[r,c].color \in \{\text{red, black, blue, green, $\emptyset$}\}$, 
$B[r,c].shape \in \{\text{star, square, triangle, circle, $\emptyset$}\}$, where we define the shape and color of an empty cell to be $\emptyset$.
$\mathcal S$ is the state space, where each state $S_t = (B_0, A_0, B_1, A_1, \cdots, B_{t-1},A_{t-1},B_t)$ is a sequence including all the historical boards $B_i$'s and actions $A_i$'s up to time step $t$. 
We define $B_{-1}$ to be an empty board before the board generation process.
$P$ is the deterministic transition probability matrix specified by the hidden rule. 
Importantly, when action $A_t$ is not accepted, resulting in no change to the board, 
we still insert the board $B_{t+1}$ and action $A_t$ to $S_{t+1}$. In that case, $B_{t+1} = B_t$.
$R$ is the deterministic reward function. 
If an action is accepted or the board is already cleared, the reward is set to be $0$; otherwise, the reward is $-1$. 
Specifically: $R(S_t, A_t) = 0$ if $B_{t+1} \neq B_t$ or $B_t = B_{-1}$; $R(S_t, A_t) = -1$, o.w.
$\mu$ is the initial state distribution. 
Since the initial state $S_0 = (B_0)$, which only includes the initial board, $\mu$ is determined by the board generating process in Section~\ref{sec:gohr}.
$H$ is the time horizon that characterizes the number time steps in each episode. 
The MDP evolves according to Section~\ref{sec:gohr}.
We define the value function as the sum of future rewards by taking action according to policy $\pi$, i.e. $V^{\pi}(S_t) = \mathbb{E}\left[\sum_{k = t}^{H-1}R(S_k, A_k)\right]$, where the expectation is over the potential randomness of the policy.
Similarly, we define the state-action value function by the sum of future rewards by taking action $A_t$ at state $S_t$ and following policy $\pi$ after that, i.e.
$Q^{\pi}(S_t,A_t) = R(S_t,A_t) + V^{\pi}(S_{t+1})$.
Because the step information has already been encoded in the state, we do not need to use step $h$ as the subscript for the value function and state-action value function.
The goal of the learner is to find a policy $\pi$ to maximize the expected cumulative reward
$\mathbb{E}_{S_0 \sim \mu}\left[V^{\pi}(S_0)\right]$. 

Even though the MDP $\mathcal{M}$ has a finite state space $\mathcal{S}$, the size of $\mathcal{S}$ scales exponentially in $H$, which makes it impractical to learn a proper policy directly.
Instead, we handcraft a feature mapping $\phi(\cdot, \cdot)$ to extract only the essential information for some particular hidden rules.
The feature function maps each state-action pair $(S_t,A)$ to a $3720$-dimensional Boolean vector $\phi(S_t,A)$.
Recall that each action $A$ chooses a cell $(r,c)$ and puts the object in this cell in bucket $b$.
$\phi(S_t,A)$ encodes the following information to Boolean variables: the shape and color information of the cell chosen by $A$, the target bucket, and the shape, color, and bucket information of the last accepted action.
Products of some of the Boolean variables are also included in $\phi(S_t,A)$.
Interested readers are referred to Appendix~\ref{sec:features} for additional details on the feature representation. 

We use a variant of DQN \cite{dqnPaper}, a model-free algorithm, as our learning algorithm.
DQN models the state-action value function $Q$ as a linear function in the feature vector $\phi$:
$Q(s,a;\theta) = \theta^\top \phi(s,a)$
and tries to estimate the optimal $Q^*$ %
using stochastic gradient updates on moving loss $L$:
\[L(\theta) = \mathbbm E\left[(y - Q(s,a;\theta))^2\right],\]
where $y = r+\max_{a'} Q(s',a';\theta^{\text{target}})$ is the target $Q$ estimate obtained for the sample $(s,a,r,s')$. 
We distinguish the running parameter iterate $\theta$ and the target parameter $\theta^{\text{target}}$.
Unlike the DQN in \cite{dqnPaper}, $\theta^{\text{target}}$ is updated less frequently to stabilize the algorithm. 

Additionally, the algorithm maintains a buffer memory of size $1000$ for past experiences and samples a batch of $128$ tuples to calculate gradients at each time step. We run the DQN with an $\epsilon$-greedy behavior policy. 
The $\epsilon$ starts at $0.9$ and terminates at $0.001$. $\epsilon$ decays exponentially with step-size $200$. 
More precisely, $\epsilon = 0.001 + (0.9-0.001)\exp(- \frac{\text{number of moves}}{200})$.
For each rule we run $100$ trials, with each trial including $200$ episodes of length $H = 100$.
The full learning algorithm is described in Appendix~\ref{sec:algorithm} and relevant hardware information is described in Appendix~\ref{sec:hardware}.

\subsection{Metrics and sample results}
\begin{figure}[h]
    \centering
    \includegraphics[width=.8\textwidth]{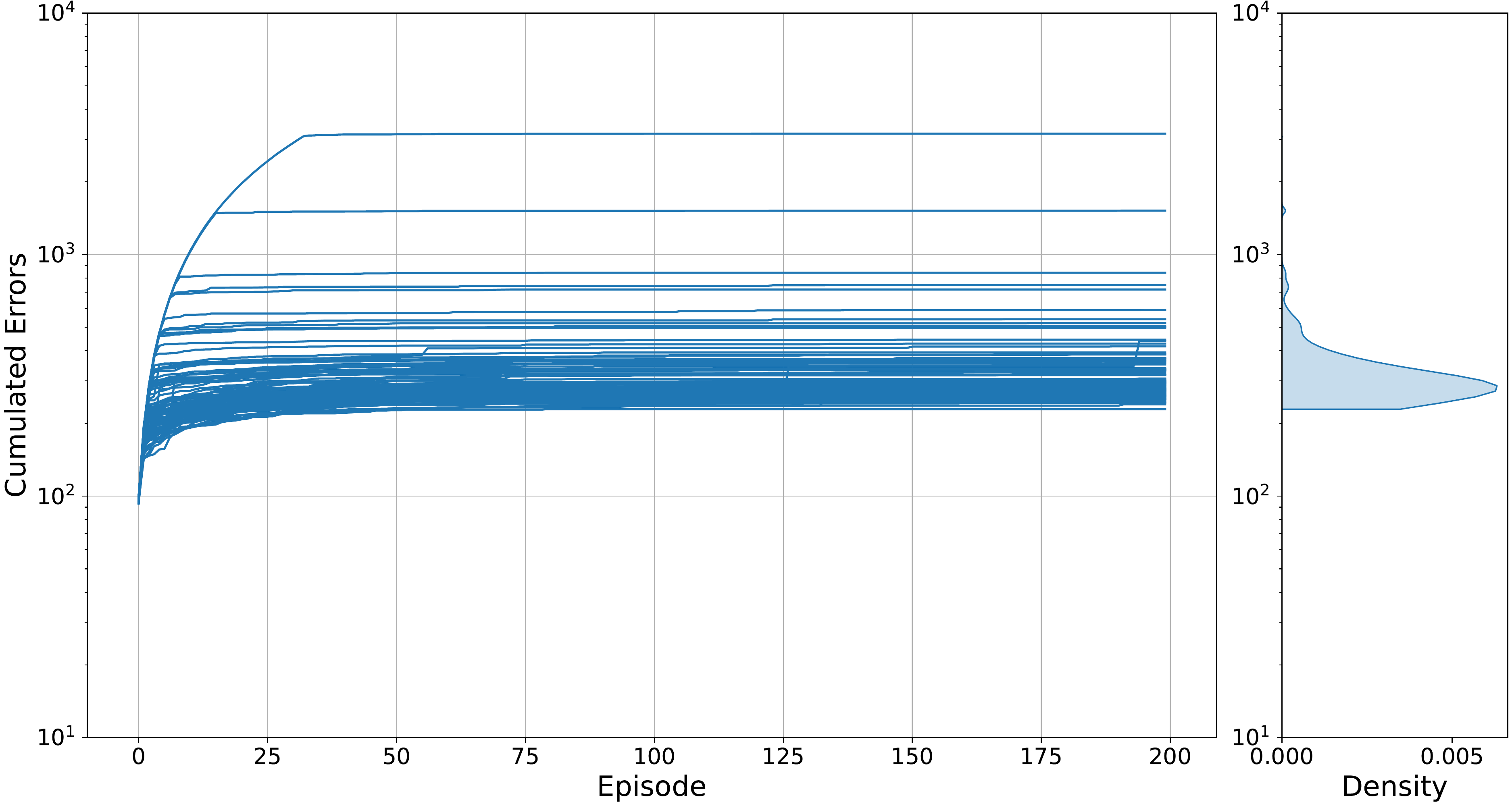}
    \caption{Cumulated error curves for 100 MLA learning runs of the Color Match rule. The y-axis is logarithmic to capture the highly skewed nature of the distribution. The marginal plot to the right displays a kernel density estimate for the Terminal Cumulated Error for this rule and MLA.}
    \label{fig:singleruleplot}
\end{figure}

\begin{figure}[h]
    \centering
    \includegraphics[width=.9\textwidth]{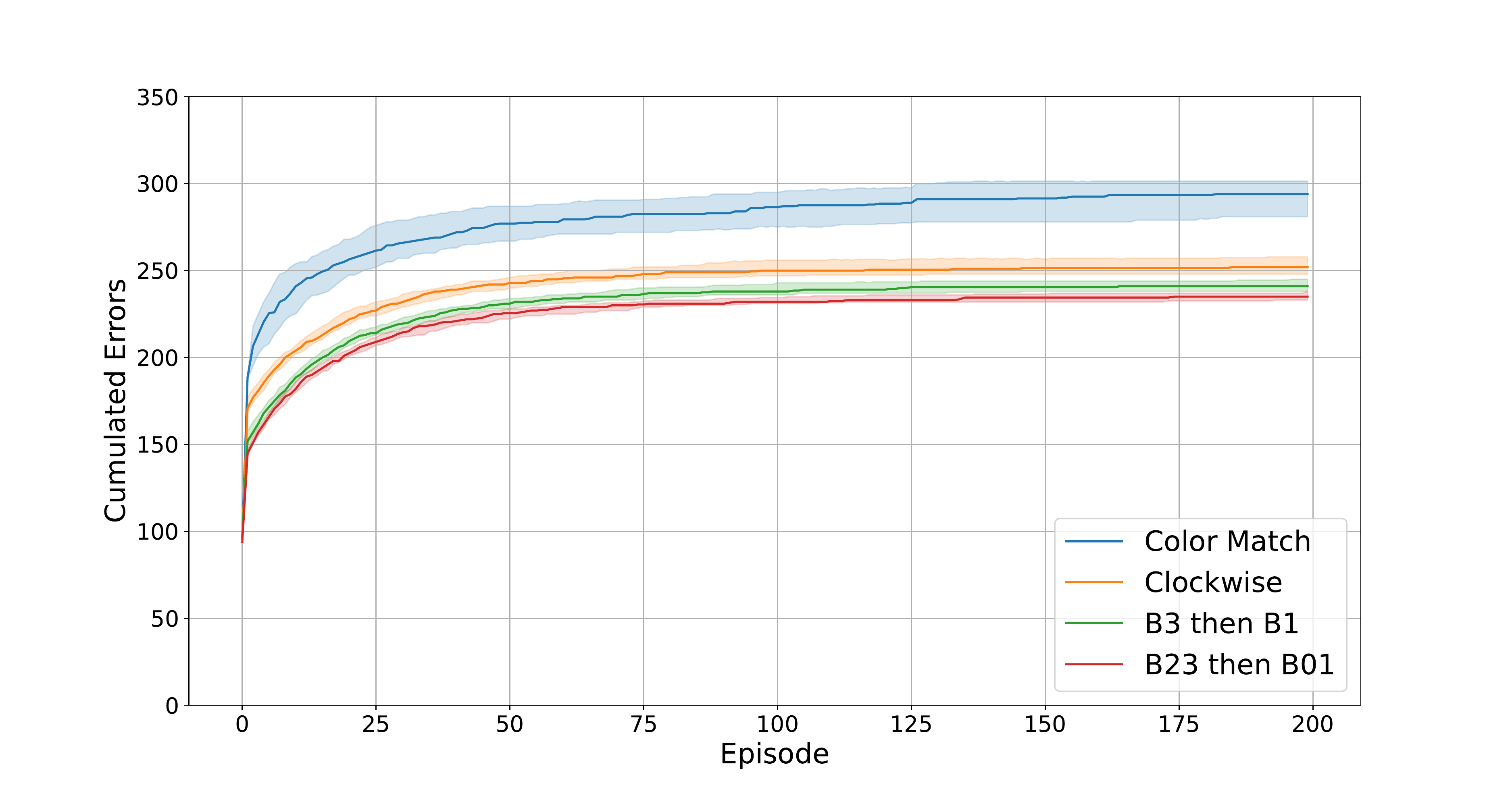}
    \caption{Median cumulated errors at each episode across 100 MLA learning runs for each sample rule. The shaded regions denote the 95\% confidence intervals for the medians, calculated using 50,000 bootstraps.}
    \label{fig:lineplot}
\end{figure}

\begin{figure}
    \centering
    \includegraphics[width=.9\textwidth]{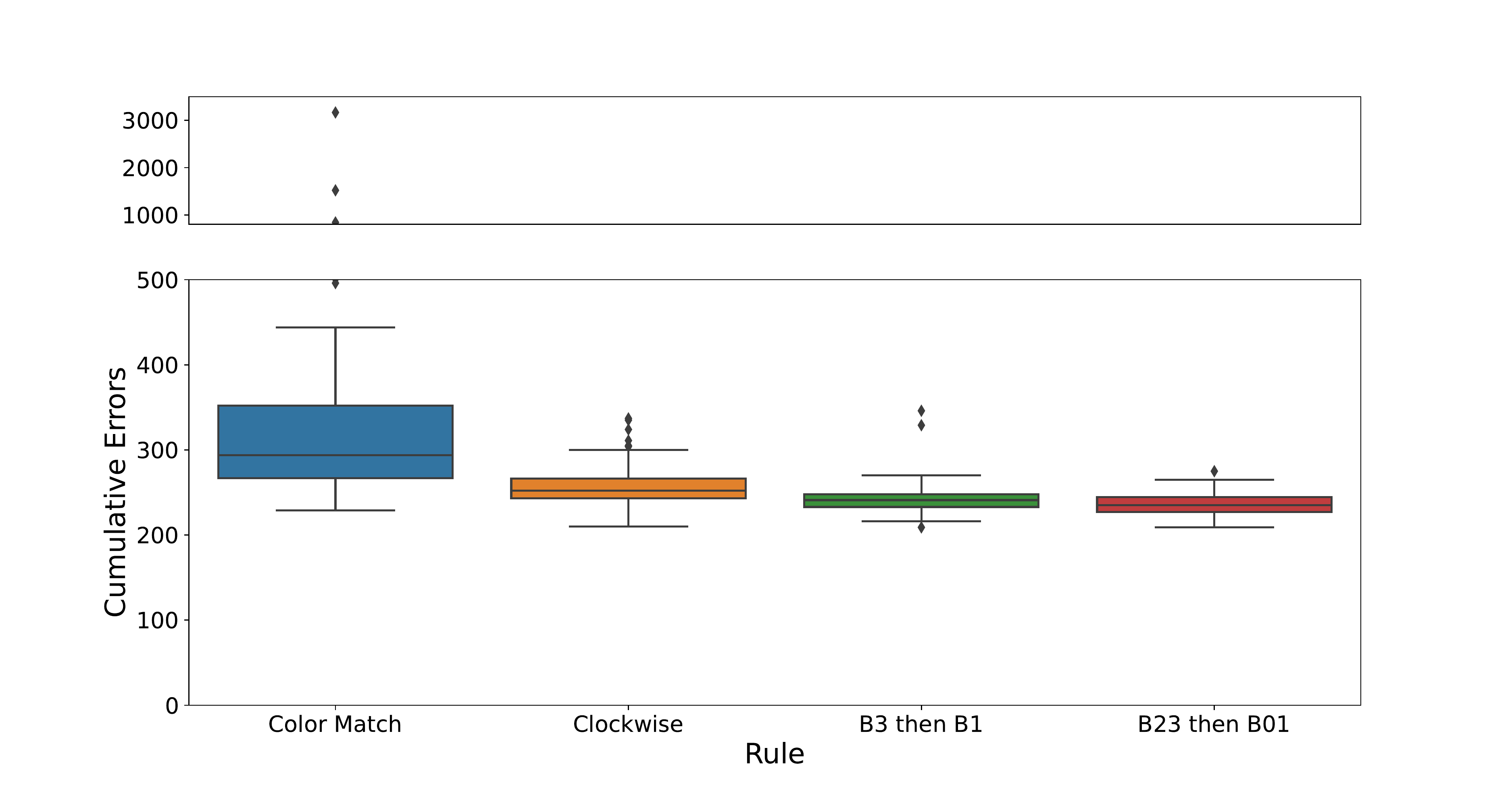}
    \caption{Box-and-whisker plots of TCE for each learning run for all sample rules. The  chart has a broken y-axis with a scale change to capture the most extreme outliers. The colored area is the interquartile range (IQR), with whiskers displaying the most extreme data point within 1.5 IQRs of the upper and lower quartiles.}
    \label{fig:brokenlinearboxplot}
\end{figure}

For any given rule and MLA, difficulty can be evaluated by examining the cumulative error count made by the MLA during each independent learning run. Given a sufficient number of episodes (200 in this analysis), we found that the cumulated error eventually flattened for all learning runs of our sample DQN; we refer to the cumulated error after 200 episodes as the Terminal Cumulated Error (TCE) for that learning run. For other rules or algorithms, longer learning runs might be necessary. As an example, Figure~\ref{fig:singleruleplot} shows the cumulated error, tallied by episode, of our sample DQN for 100 separate learning runs of the Color Match rule. Median learning curves, also plotted by episode, across all four sample rules can be found in Figure~\ref{fig:lineplot}. We found median cumulated error plots to be a useful tool for visualizing differences in difficulty among a set of rules, with higher curves indicating greater difficulty. Box-and-whisker plots for the distributions of TCE for each sample rule are shown in Figure~\ref{fig:brokenlinearboxplot}, with higher distributions again indicating greater difficulty.


Although the heavily-skewed TCE distribution seen for Color Match was not seen for all sample rules, it highlights the need for non-parametric tests when comparing rules. To perform pairwise comparison of rule difficulty, for instance between arbitrary rules $A$ and $B$, we recommend using the Mann-Whitney-Wilcoxon U-Test \citep{nachar_mann-whitney_2008} with TCE as the point-metric. The test compares every TCE value associated with rule $A$ against every TCE value associated with rule $B$; the associated test statistic sums the number of times that values from $A$ exceed values from $B$, with ties counting for 0.5. We found that one-sided U-Tests ranked rules identically ($p<0.002$) to the difficulty order given by the height of the median cumulated error curves in Figure~\ref{fig:lineplot}. The sample rules were chosen to illustrate a range of rules possible within the rule game, not with a particular variation in task characteristics in mind; we expect interested researchers to design targeted sets of rules investigating certain characteristics of learning tasks and to perform similar comparisons to assess their practical difficulty.

In addition to comparing the difficulty of two different rules for a given algorithm, the U-Test also supports comparison between different MLAs. The U-test test statistics can be used to form an ``ease ratio,'' representing the ratio of learning runs that, for example, rule $A$ was found easier than rule $B$ by a given MLA. We propose aggregating ease ratios for many MLAs across the set of benchmark rules to create a tournament graph as described in \citep{strang_network_2022}. The ranking method corrects for the fact that algorithms $R$ and $S$ may have solved a different subset of rules, leading to the possibility that $R$ encountered a larger fraction of ``easy'' rules (where rule difficulty can be assessed by the TCE of all algorithms that attempt a particular rule). Additional information regarding the methods to be used in ranking competition submissions can be found at the \href{http://sapir.psych.wisc.edu:7150/w2020/captive.html}{public site}.

\section{Discussion}
\label{sec:discussion}

The GOHR tool provides a capability for studying the performance of MLAs in a novel and principled way. Using the expressive rule syntax, researchers can make precise changes to rules of interest to study how they affect algorithm performance.
As suggested in Section~\ref{sec:sample}, this approach can be used to compare the performance of different algorithms on any specific set of rules and to see the effect of rule characteristics (position, properties, static/dynamic) 
on MLA performance. This is a useful complement to present competitions among ML algorithms on existing board games and video games that can only look at complex tasks without the same granularity.  We note here that the GOHR can also be played by human rule learners, enabling comparison of human learning and ML on an equal footing. For additional discussion on this  please refer to \cite{bier_can_2019}.

Each choice of ML parametrization, learning model, and optimization is expected to produce different results. With the GOHR, we believe it will be possible to interrogate how these choices interact with specific rule characteristics to result in differences in practical performance.
With this goal in mind, we are sharing the complete suite of tools with all interested researchers. 
We hope that researchers using the GOHR tool for ML research will share their findings to help this inquiry. 

Given our current analysis and discussion, we believe there are at least two interesting broad research questions that can be answered with the GOHR environment. First, can a given ML algorithm learn many substantially different rules, thus approaching a general learning. Second, when results for multiple algorithms are available, is it possible to cluster rules into difficulty classes based on practical algorithm performance.
\begin{ack}
Support for this research was provided by the University of Wisconsin-Madison Office of the Vice Chancellor for Research and Graduate Education with funding from the Wisconsin Alumni Research Foundation, and by the National Science Foundation under Grant No. 2041428. Any opinions, findings, and conclusions or recommendations expressed in this material are those of the author(s) and do not necessarily reflect the views of the sponsors. We also thank Mr. Kevin Mui for building the first instance of the GOHR, to support research on human learning. 
\end{ack}

\bibliographystyle{plainnat}
\bibliography{neurips_2022_dataset}

\appendix
\newpage
\section{Appendix}
\subsection{Details of the reinforcement learning algorithm}
\subsubsection{Feature representation\label{sec:features}}
The feature vector $\phi(S,A)$ is constructed with the following four different types of components:
\begin{enumerate}[leftmargin=*,itemsep=0pt]
	\item The 12 current-step unary features $\phi_u(S,A)$, where $u$ can be a color, a shape, or a bucket.
If $u$ is a color: $$\phi_u(S_t,A) = \id{\{B_t[A.r, A.c].\text{color} = u\}}.$$
If $u$ is a shape: $$\phi_u(S_t,A) = \id{\{B_t[A.r, A.c].\text{shape} = u\}}.$$
If $u$ is a bucket: $$\phi_u(S_t,A) = \id{\{A.b = u\}}.$$

	\item The 48 current-step binary features, a set of binary features that depend on the current board configuration $\phi_{u,v} : \mathcal S \times \mathcal A \rightarrow \{0,1\}$ where $(u,v)$ can be a (color, shape), (color, bucket) or (shape, bucket) tuple. These features $\phi_{u,v}(S_t, A)$ are defined as follows:
	
	\renewcommand{\arraystretch}{2}
	\begin{center}
		\begin{tabular}{c|c}
			$u,v \in $ & $\phi_{u,v}(S_t, A)$\\
			\hline
			color $\times$ shape & $\id{\{B_t[A.r, A.c].\text{color} = u\}} \wedge \id{\{B_t[A.r, A.c].\text{shape} = v\}}$\\
			\hline
			color $\times$ bucket & $\id{\{B_t[A.r, A.c].\text{color} = u\}} \wedge \id{\{A.b = v\}}$\\
			\hline
			shape $\times$ bucket & $\id{\{B_t[A.r, A.c].\text{shape} = u\}} \wedge \id{\{A.b = v\}}$\\
		\end{tabular}
	\end{center}
	
	\item The 60 last-successful-step and current-step binary features, a set of binary features constructed from the board and action information at the last successful step and the current step.  For a time step $t>0$, let $$\ell(t) = \max\{ \tau \in \{-1, 0,\cdots, t-1\} \text{ s.t. } B_t \neq B_{\tau}\}$$ be the latest successful time step. Since $B_{-1}$ is an empty board, $B_{-1} \neq B_t,\, \forall t \geq 0$ until the game is over, so $\ell(t)$ is well defined.

	The features are defined as follows: $\phi_{(u),v} : \mathcal S \times \mathcal A \rightarrow \{0,1\}$, where $(u)$ is extracted from the board and action information at the last successful time step, and $v$ is extracted from the board and action information at the current time step. Note that $u$ can be the empty set $\emptyset$ when $\ell(t) =-1$.
	
	\begin{center}
		\begin{tabular}{c|c}
			$(u),v \in $ & $\phi_{(u),v}(S_t, A)$\\
			\hline
			(color$^{\emptyset}$) $\times$ color & $\id{\{B_{\ell(t)}[A_{\ell(t)}.r, A_{\ell(t)}.c].\text{color} = u\}} \wedge \id{\{B_t[A.r, A.c].\text{color} = v\}}$\\
			\hline
			(shape$^{\emptyset}$) $\times$ shape & $\id{\{B_{\ell(t)}[A_{\ell(t)}.r, A_{\ell(t)}.c].\text{shape} = u\}} \wedge \id{\{B_t[A.r, A.c].\text{shape} = v\}}$\\
			\hline
			(bucket$^{\emptyset}$) $\times$ bucket & $\id{\{A_{\ell(t)}.b = u\}} \wedge \id{\{A.b = v\}}$\\
		\end{tabular}
	\end{center}
	where $x^\emptyset = x \cup \{\emptyset\},\, \forall x \in \text{\{color, shape, bucket\}}$

	\item The 3600 last-successful-step and current-step 4-ary features, a set of 4-ary features constructed from the board and action information at the last successful step and the current step. $$\phi_{(u, v), y,z} : \mathcal S \times \mathcal A  \rightarrow \{0,1\}$$
	where $(u, v)$ are extracted from the board and action information at the last successful time step, and $y, z$ are extracted from the current time step. These features, $\phi_{(u,v),y,z}(S_t, A)$, are defined as follows:
	
		\begin{center}
		\begin{tabular}{c|c}
			$(u,v), y,z \in $ & $\phi_{(u,v),y,z}(S_t, A)$\\
			\hline
			$\begin{aligned}
			&(\she,\coe) \\
			\x & \sh, \co
			\end{aligned}$
			&
			$\begin{aligned}
			&\id{\{B_{\ell(t)}[A_{\ell(t)}.r, A_{\ell(t)}.c].\text{shape} = u\}}\\ \wedge & \id{\{B_{\ell(t)}[A_{\ell(t)}.r, A_{\ell(t)}.c].\text{color} = v\}}\\
			\wedge&\id{\{B_t[A.r, A.c].\text{shape} = y\}} 
			\wedge\id{\{B_t[A.r, A.c].\text{color} = z\}}
			\end{aligned}$\\
            $\begin{aligned}
			&(\she,\coe) \\
			\x & \sh, \bu
			\end{aligned}$
			&
			$\begin{aligned}
			&\id{\{B_{\ell(t)}[A_{\ell(t)}.r, A_{\ell(t)}.c].\text{shape} = u\}}\\ \wedge & \id{\{B_{\ell(t)}[A_{\ell(t)}.r, A_{\ell(t)}.c].\text{color} = v\}} \\
			\wedge & \id{\{B_t[A.r, A.c].\text{shape} = y\}} \wedge \id{\{A.b = z\}}
			\end{aligned}$ \\
            $\begin{aligned}
			&(\she,\coe) \\
			\x & \co, \bu
			\end{aligned}$
			&			
			$\begin{aligned}
			&\id{\{B_{\ell(t)}[A_{\ell(t)}.r, A_{\ell(t)}.c].\text{shape} = u\}}\\ 
			\wedge &  \id{\{B_{\ell(t)}[A_{\ell(t)}.r, A_{\ell(t)}.c].\text{color} = v\}} \\
			\wedge & \id{\{B_t[A.r, A.c].\text{color} = v\}} \wedge \id{\{A.b = z\}}
			\end{aligned}$\\
			\hline
            $\begin{aligned}
			&(\she,\bue) \\
			\x & \sh, \co
			\end{aligned}$
			&
			$\begin{aligned}
			&\id{\{B_{\ell(t)}[A_{\ell(t)}.r, A_{\ell(t)}.c].\text{shape} = u\}} \wedge  \id{\{A_{\ell(t)}.b = v\}}\\
			\wedge &\id{\{B_t[A.r, A.c].\text{shape} = y\}} 
			\wedge  \id{\{B_t[A.r, A.c].\text{color} = z\}}
			\end{aligned}$\\
            $\begin{aligned}
			&(\she,\bue) \\
			\x & \sh, \bu
			\end{aligned}$
			&			
			$\begin{aligned}
			&\id{\{B_{\ell(t)}[A_{\ell(t)}.r, A_{\ell(t)}.c].\text{shape} = u\}} \wedge \id{\{A_{\ell(t)}.b = v\}}\\
			\wedge &\id{\{B_t[A.r, A.c].\text{shape} = y\}}
			\wedge \id{\{A.b = z\}}
			\end{aligned}$ \\
            $\begin{aligned}
			&(\she,\bue) \\
			\x & \co, \bu
			\end{aligned}$
			&			
			$\begin{aligned}
			&\id{\{B_{\ell(t)}[A_{\ell(t)}.r, A_{\ell(t)}.c].\text{shape} = u\}} 
			\wedge  \id{\{A_{\ell(t)}.b = v\}} \\
			\wedge & \id{\{B_t[A.r, A.c].\text{color} = y\}} 
			\wedge \id{\{A.b = z\}}
			\end{aligned}$\\
			\hline
            $\begin{aligned}
			&(\coe,\bue) \\
			\x & \sh, \co
			\end{aligned}$
			&			
			$\begin{aligned}
			&\id{\{B_{\ell(t)}[A_{\ell(t)}.r, A_{\ell(t)}.c].\text{color} = u\}} \wedge \id{\{A_{\ell(t)}.b = v\}} \wedge\\
			& {\id{\{B_t[A.r, A.c].\text{shape} = y\}} \wedge \id{\{B_t[A.r, A.c].\text{color} = z\}}}
			\end{aligned}$\\
            $\begin{aligned}
			&(\coe,\bue) \\
			\x & \sh, \bu
			\end{aligned}$
			&
			$\begin{aligned}
			&\id{\{B_{\ell(t)}[A_{\ell(t)}.r, A_{\ell(t)}.c].\text{color} = u\}} \wedge \id{\{A_{\ell(t)}.b = v\}} \\
			\wedge & {\id{\{B_t[A.r, A.c].\text{shape} = y\}} \wedge \id{\{A.b = z\}}}
			\end{aligned}$ \\
            $\begin{aligned}
			&(\coe,\bue) \\
			\x & \co, \bu
			\end{aligned}$
			&			
			$\begin{aligned}
			&\id{\{B_{\ell(t)}[A_{\ell(t)}.r, A_{\ell(t)}.c].\text{color} = u\}} \wedge \id{\{A_{\ell(t)}.b = v\}} \\
			\wedge
			&{\id{\{B_t[A.r, A.c].\text{color} = y\}} \wedge \id{\{A.b = z\}}}
			\end{aligned}$
			\\
			\hline
		\end{tabular}
	\end{center}
\end{enumerate}
For any $(S, A)$-tuple, the final feature vector $\phi(S,A)$ is created by stacking up all of the above features, which gives a feature function $\phi:\mathcal S \times \mathcal A \rightarrow \{0,1\}^{3720}$.

\subsubsection{Learning algorithm} \label{sec:algorithm}

\floatname{algorithm}{Algorithm}
\begin{algorithm}[htbp!]
	
	\begin{flushleft}
		\textbf{Input} : target iteration $T$, episode count $M$, buffer size $N$, batch size $B$.
	\end{flushleft}

	\begin{algorithmic}
		\State Initialize experience replay buffer memory $\mathcal{D}$ with capacity $N$.
		\State Initialize the iterate parameters  $\theta^{\text{target}}_1 \leftarrow 0,\theta_{1,1} \leftarrow 0$.
		\For{target iteration $\tau = 1,\cdots,T$}
		\For{episode $e = 1,\cdots,M$} 
		\State Draw $s_1 \sim \mu$.
		\For {time step $t=1,\cdots,H$} 
		\State Draw $a_t \sim \epsilon \cdot \text{Unif}(A) + (1-\epsilon)\cdot  \mathbbm{1}_A\{a = \max_{a'} Q(s_t, a'; \theta_{e,t})\}$.
		\State Take action $a_t$ and observe reward $r_t$ and observation $x_{t+1}$.
		\State Set $s_{t+1} \leftarrow (s_t, a_t, x_{t+1})$ and enque $\left(s_t,a_t,r_t,s_{t+1}\right)$ to $\mathcal D$.
		\State Uniformly sample a batch $\{\left(s_j,a_j,r_j,s'_{j}\right)\}|_{j=1}^B$ from $\mathcal{D}$ and 
		\State \quad set $\forall j \in [B]$, $y_j = r_j + \mathbbm{1}\{j \text{ is terminal}\}\cdot \gamma \max_{a'} Q(s'_{j}, a'; \theta^{\text{target}}_\tau)$
		\State Update $\theta_{e,t+1}$ using $Q$-objective $\frac{1}{B}\sum_{j=1}^{B}  \left(y_j - Q(s_j,a_j ; \theta ) \right)^2\big|_{\theta = \theta_{e,t}}.$
		\EndFor
		\State Update the parameter for next episode $\theta_{e+1, 1} \leftarrow \theta_{e,H+1}$.
		\EndFor
		\State Update the target parameter $\theta^{\text{target}}_{\tau+1} \leftarrow \theta_{M,H+1}$.
		\EndFor
	\end{algorithmic}
	\caption{Deep Q-learning(DQN) with Experience Replay}
	\label{alg}
\end{algorithm}

\subsubsection{Training hardware} \label{sec:hardware}
For all experiments, we used a local server (at the University of Wisconsin - Madison) running Ubuntu 20.04 LTS, with 2 Intel Xeon Silver 4214R CPUs (2.40 GHz) and 196GB of RAM. No GPUs were used in our experiments.

\subsection{Code Licensing}
The source code for the GOHR, provided at our public site, is licensed under the Apache License version 2.0.

\end{document}